\def\year{2019}\relax
\documentclass[letterpaper]{article} 
\usepackage{aaai19}  
\usepackage{times}  
\usepackage{helvet} 
\usepackage{courier}  
\usepackage[hyphens]{url}  
\usepackage{graphicx} 
\def\UrlFont{\rm}  
\usepackage{graphicx}  
\usepackage{amsmath}

\usepackage[normalem]{ulem}
\usepackage{multirow}
\usepackage{booktabs}

\def\etal{\emph{et al.}}
\usepackage{algorithm}
\usepackage[noend]{algpseudocode}
\usepackage{amsfonts}
\usepackage{color}

\usepackage[switch]{lineno}

\title{Learning Graph Convolutional Network for Skeleton-based Human Action Recognition by Neural Searching}
\author{Wei Peng\textsuperscript{1}, Xiaopeng Hong\textsuperscript{2,1}, Haoyu Chen\textsuperscript{1}, Guoying Zhao\textsuperscript{1,}\thanks{Corresponding author}\\
\textsuperscript{1}Center for Machine Vision and Signal Analysis, University of Oulu, Finland\\
\textsuperscript{2}Faculty of Electronic and Information Engineering, Xi'an Jiaotong University, PRC \\
{\tt\small \{wei.peng, xiaopeng.hong, chen.haoyu, guoying.zhao\}@oulu.fi}
}

\begin{document}

\maketitle

\begin{abstract}
Human action recognition from skeleton data, fueled by the Graph Convolutional Network (GCN), has attracted lots of attention, due to its powerful capability of modeling non-Euclidean structure data. However, many existing GCN methods provide a pre-defined graph and fix it through the entire network, which can loss implicit joint correlations. Besides, the mainstream spectral GCN is approximated by one-order hop, thus higher-order connections are not well involved. Therefore, huge efforts are required to explore a better GCN architecture. To address these problems, we turn to Neural Architecture Search (NAS) and propose the first automatically designed GCN for skeleton-based action recognition. Specifically, we enrich the search space by providing multiple dynamic graph modules after fully exploring the spatial-temporal correlations between nodes. Besides, we introduce multiple-hop modules and expect to break the limitation of representational capacity caused by one-order approximation. Moreover, a sampling- and memory-efficient evolution strategy is proposed to search an optimal architecture for this task. The resulted architecture proves the effectiveness of the higher-order approximation and the dynamic graph modeling mechanism with temporal interactions, which is barely discussed before. To evaluate the performance of the searched model, we conduct extensive experiments on two very large scaled datasets and the results show that our model gets the state-of-the-art results. 


\end{abstract}

\section{Introduction}

Human action recognition is a valuable but challenging research area with widespread potential applications, say security surveillance, human computer interaction and autonomous driving. Nowadays, as an alternative to the appearance and depth data, skeleton data is popularly used in action recognition. One important reason is that skeleton data conveys compact information of body movement, thus it is robust to the complex circumstances like the variations of the viewpoints, occlusion and self-occlusion. Previous works reorganize the skeleton data into a kind of grid-shape structure so that the traditional recurrent neural networks (RNN) and convolutional neural networks (CNN) can be implemented. Though substantial improvements have been seen in action recognition, it does not fully benefit from the superior representing capability of deep learning, as the skeleton data lies in a non-Euclidean geometric space. Currently, Graph Convolutional Networks (GCN)~\cite{kipf2016semi,defferrard2016convolutional} has been introduced to skeleton-based action recognition and achieved many encouraging results~\cite{li2018spatio,yan2018stgan,li2019spatio,gao2019optimized,shi2019two,li2019actional}. Nonetheless, most GCN methods are based on a pre-defined graph with fixed topology constraint, which ignores implicit joint correlations. Work in~\cite{shi2019two} turns to replace the fixed graph with an adaptive one based on the node similarity. However, it provides a shared mechanism through the entire network and the spatial-temporal correlations are barely discussed. We argue that different layers contain different semantic information thus a layer-specific mechanism should be involved to construct a dynamic graph. Besides, mainstream GCN tends to one-order Chebyshev polynomials approximation~\cite{kipf2016semi} to reduce the computational expense, meanwhile high-order connections are not well involved so that the representational ability is limited. Current works, like~\cite{gao2019optimized}, introduce high-order approximation to have GCN with a bigger receptive filed. Nonetheless, the contribution of the each component in the approximation is not discussed. {It is apparent that designing such different function modules for different tasks requires lots of efforts and exhausted try-and-error tests.}\par
To address this problem, in this paper, we focus on reducing the manual efforts in designing better graph convolutional architecture. We replace the fixed graph structure with dynamic ones by Automatic Neural Architecture Search (NAS)~\cite{zoph2016neural} and explore different graph generating mechanisms at different semantic levels. NAS is designed to obtain superior neural network structures with less or without human assists under a reasonable computational budgets. However, it is not straightforward to apply NAS to GCN. Graph data like skeleton has no locality and order information as required by convolution operations, while current NAS methods focus on the design of neural operations. Besides, GCN itself is a relative new research area thus existing operations are very limited, e.g., GCN does not even have a general pooling operation. Therefore, we propose to search in a GCN space built with multiple graph function modules. Moreover, a high sample-efficient deep neuro-evolution strategy (ES)~\cite{angeline1994evolutionary,miller1989designing} is provided to explore an optimal GCN architecture by estimating the architecture distribution. It could be conducted in both continuous and discrete search space. Thus, one can just activate one function module at each iteration to search by a memory-efficient fashion. With our NAS for GCN, we automatically build a graph convolutional network for action recognition from skeleton data. {To evaluate the proposed method, we perform comprehensive experiments on two large-scaled datasets, NTU RGB+D~\cite{shahroudy2016ntu} and Kinetcis-Skeleton~\cite{kay2017kinetics,yan2018stgan}. Results show that our model is robust to the subject and view variations and achieves the state-of-the-art performance.} The contributions of this paper are manifold:
\begin{itemize}
    \item {We break the limitation of GCN caused by its fixed graph and, for the first time, determine the graph convolution architecture with NAS for skeleton-based action recognition.}
    \item {We enrich the search space for GCN from the following two aspects. Firstly, we provide multiple dynamic graph substructures on the basis of various spatial-temporal graphs modules. Secondly, we enlarge the receptive field of GCN convolution by building higher-order connections with Chebyshev polynomial approximation.}  
    \item {To improve the search efficiency, we devise a novel evolution based NAS search strategy, which is both sampling- and memory-efficient.}    
    

\end{itemize}

\section{Related work}

\textbf{Skeleton-based Action Recognition}
In human action recognition, as an alternative data source for RGB and depth data, skeleton data is increasingly attracted attention thanks to its robustness property against changes in body scales, viewpoints and backgrounds. Different from the grid data, the graph constructed by the human skeleton lies in a non-Euclidean space. To benefit from the great representation ability of deep learning, conventional methods tend to rearrange the skeleton data into grid-shape structure and feed it directly into the classical RNN~\cite{shahroudy2016ntu,song2017end,zhang2017view} or CNN~\cite{kim2017interpretable,liu2017enhanced} architectures. However, as mentioned in~\cite{monti2017geometric}, one can not express a meaningful operator in the vertex domain. Therefore, current works tend to GCN and prefer to build operators in the non-Euclidean space. Yan~\etal ~and Li~\etal~are the first to use GCN for skeleton-based action recognition~\cite{yan2018stgan,li2018spatio}. Gao~\etal~proposed a sparsified graph regression based GCN~\cite{gao2019optimized} to exploit the dependencies of each joints. Shi~\etal~gave a two-stream GCN architecture, in which the joints and the second-order information (bones) are both used. With a score-level fusion strategy, it gets the current best result~\cite{shi2019two}. Our method is also based on GCN and we will fully explore the influence of the graph topology for this task.

\noindent \textbf{Neural Architecture Search}
As an important part of automated machine learning (AutoML), Neural Architecture Search (NAS)~\cite{zoph2016neural} is to automatically build a neural network under a low cost of computational resources. Numerous approaches for NAS already exist in the literature, including black-box optimization based on reinforcement learning~\cite{zoph2016neural}, evolutionary search~\cite{real2018regularized}, and gradient-based method~\cite{liu2018darts}. Besides, promising progresses are also seen in aspects such as searching space design~\cite{liu2018darts}, and architecture performance evaluation~\cite{saxena2016convolutional,real2018regularized}. Automatically designed architectures have already got superior performances against the famous manual ones in the fields like image classification tasks~\cite{zoph2018learning}, and semantic image segmentation~\cite{liu2019auto}. There are also some attempts about NAS on action recognition~\cite{peng2019video} from RGB data. However, little NAS-based methods providing a solution to the non-Euclidean data. In fact, currently Gao~\etal~transferred ENAS~\cite{pham2018efficient} to graph neural network for citation networks and inductive learning tasks. However, compare to our task, it is totally different since it aims to find the transforming, propagating and aggregating functions for the network with only two or three layers.    

\noindent\textbf{GCN and Attention mechanism}  Graph neural network is widely used on irregular data like social networks, and biological data. Generally, there are two ways to define a GCN. The spectral-domain method~\cite{defferrard2016convolutional,kipf2016semi} models the representation in the graph Fourier transform domain based on eigen-decomposition, meanwhile it is time-consuming. The Nodal-domain method~\cite{monti2017geometric,velivckovic2018graph} directly implements GCN on the graph node and its neighbors. However, it is difficult to model the global structure. To further improve the performance of GCN, attention mechanisms is introduced to GCN~\cite{velivckovic2018graph,vaswani2017attention}. One benefit of attention mechanisms is that they select information which is relatively critical from all inputs. Inspired by this, Velickovic~\etal leveraged attention mechanism for graph node classification and achieved state-of-the-art performance~\cite{velivckovic2018graph}. Work in~\cite{sankar2018dynamic} employs self-attention along both spatial and temporal dimensions and get superior results on link prediction tasks. Nonetheless, our work is different since we compute the interaction between nodes based on various semantic information to build a dynamic graph, while others is to compute the importance weights either for frames or different feature representations.


\section{Methodology}
In this section, we detail our search-based GCN for action recognition from skeleton data. To make the paper self-contained, we briefly review how to model a spatial graph with GCN first.

Consider an undirected graph $\mathcal{G} = \{\mathcal{V}, \mathcal{E}, A\}$ composed of $n=|\mathcal{V}|$ nodes, which are connected by $|\mathcal{E}|$ edges and the node connections are encoded in the adjacency matrix $A \in \mathcal{R}^{n \times n}$. Let $X \in \mathcal{R}^{n}$ be the input representation of $\mathcal{G}$ and$\{ x_i, \forall i \in \mathcal{V}\}$ be its $n$ elements. Then to model the representation of $\mathcal{G}$, the graph is performed a graph Fourier transform so that the transformed signal, as in the Euclidiean space, could then perform formulation of fundamental operations such as filtering. Therefore, the graph Laplacian $L$, of which the normalized definition is $L = I_n - D^{-1/2}AD^{-1/2}$ and $D_{ii} = \sum_{j}A_{ij}$, is used for Fourier transform. Then a graph filtered by operator $g_{\theta}$, parameterized by $\theta$, can be formulated as
\begin{equation}
    Y = g_{\theta}(L)X = U g_{\theta}(\Lambda) U^{T}X,
\end{equation}
where $Y$ is the extracted feature of node. $U$ is the Fourier basis and it is a set of orthonormal eigenvectors for $L$ so that $L = U \Lambda U^T$ with the $\Lambda$ as its corresponding eigenvalues. However, multiplication with the eigenvectors matrix is expensive. The computational burden of this non-parametric filter is $\mathcal{O}(n^2)$~\cite{defferrard2016convolutional}. Suggested by~\cite{hammond2011wavelets}, the filter $g_{\theta}$ can be well-approximated by a Chebyshev polynomials with $R$-th order.
\begin{equation}\label{eq:cheb_approximation}
    Y = \sum_{r=0}^{R}\theta_r^{'}T_r(\hat{L})X,
\end{equation}

\noindent of which $\theta_r^{'}$ denotes Chebyshev coefficients. Chebyshev polynomial $T_r(\hat{L})$ is recursively defined as
\begin{equation}\label{eq:cheb}
    T_r(\hat{L}) = 2\hat{L}T_{r-1}(\hat{L})-T_{r-2}(\hat{L})
\end{equation}
 with $T_0 = 1$ and $T_1 = \hat{L}$. Here $\hat{L} = 2L/{\lambda_{max}}-I_n$ is normalized to [-1,1]. For Eq~(\ref{eq:cheb_approximation}), work in~\cite{kipf2016semi} sets $R=1$, $ \lambda_{max}= 2$ and makes the network adapt to this change. In this way, a first-order approximation of spectral graph convolutions is formed. Therefore, 
\begin{equation}
    Y = \theta_0^{'}X +\theta_1^{'}(L-I_n)X = \theta_0^{'}X -\theta_1^{'}(D^{-1/2}AD^{-1/2})X.
\end{equation}

\noindent Likewise, $\theta_r^{'}$ can also be approximated with an unified parameter $\theta$, which means $\theta = \theta_0 = - \theta_1$, and let the training process adapt the approximation error, then 
\begin{equation}
    Y = \theta(I_n + D^{-1/2}AD^{-1/2})X.
\end{equation}
The computational expense is $\mathcal{O}(|\mathcal{E}|)$. One can stack multiple GCN layers to get high-level graph feature. To make it simple, in the following sections, we set  $ L = I_n + D^{-1/2}AD^{-1/2}$, and generally, $X\in \mathcal{R}^{n\times C}$ is with multi-channels. Thus 
\begin{equation}
    Y = LX\theta.
\end{equation}

\subsection{Searched Graph Convolutional Network}
Here we consider the human action recognition problem from skeleton data as a graph classification task from a sequence of graphs $\mathbb{G} = \{\mathcal{G}_{1}, \mathcal{G}_{2},...,\mathcal{G}_{T}\}$. Each graph denotes a skeleton at a certain time step and its nodes and edges represent the skeleton joints and bones, respectively. Then, this task can be framed as supervised learning problem on graph data, in which the goal is to learn a robust representation of $\mathbb{G}$ with GCN and thus to give a better prediction of action classes. To this end, we propose to construct this GCN with neural architecture search, which automatically assembles graph generating modules for layers at different semantic levels. Firstly, we will detail the GCN search space built with different graph modules. Then, we present a sampling- and memory-efficient search strategy.

\subsubsection{GCN search space}
In NAS, a neural search space determines what and how neural operations a searching strategy could take to build a neural network. Here we search in space built with multiple GCN modules to explore the optimal module combination for dynamic graph at different representation levels. Work in~\cite{yan2018stgan} presents a ST-GCN block, which takes skeleton data and a fixed graph as inputs, to extract the spatial-temporal representation of nodes. Our GCN-block is also a spatial-temporal block, while instead of providing a pre-defined graph, we generate dynamic graphs based on the node correlations captured by different function modules. There are mainly two kinds of correlations being captured to construct the dynamic graph.

\begin{figure}[thbp!]
\centering
\includegraphics[width=0.5\textwidth]{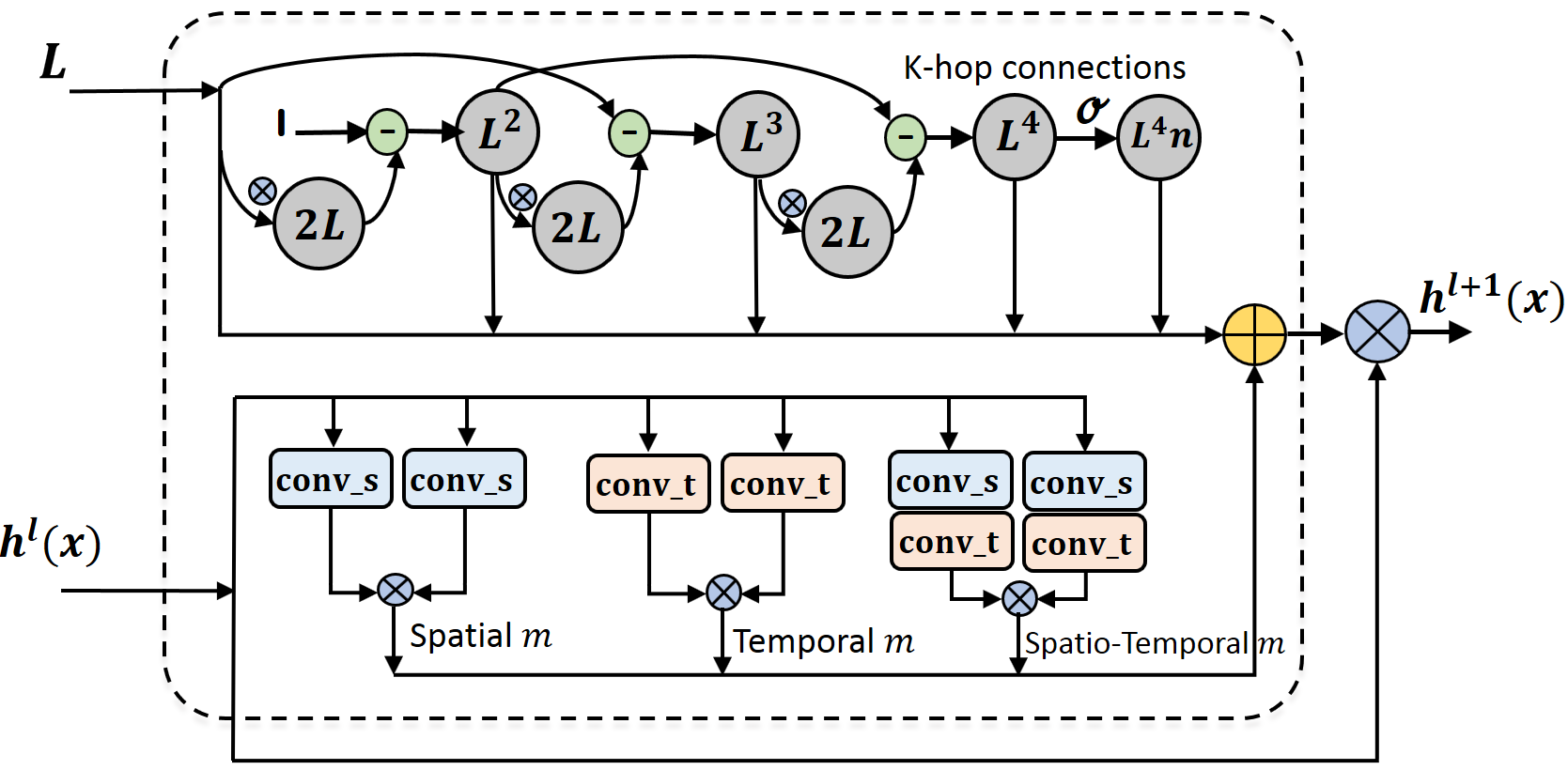}
\caption{\small{ Illustration of the search space. Here, $\bigotimes$ denotes matrix multiplication. $\bigoplus$ is the element-wise summation. There are eight function modules for generating graphs. The top part is a implementation of Chebyshev polynomial based on Eq~(\ref{eq:cheb}). We also add its separate components to the graphs and let the network choose the final ones. The bottom part contains three dynamic graph modules. All the graphs are added together according to Eq~(\ref{eq:weighted}). The contribution of each module works as the architecture parameters. Note that there is a softmax function before the summation operation for dynamic graphs. 
}}
\label{Fig:framework}
\vspace{0.5em}
\end{figure}

\noindent \textbf{Structure representation Correlation.} Structure correlation is computed based on the spatial node connections. To determine how strong the connection is between two nodes, like in~\cite{shi2019two}, a normalized Gaussian function is applied on the graph nodes and the similarity score works as the correlation. That is
\begin{equation}\label{eq:gaussain}
   \forall i,j \in \mathcal{V}, A_{D}(i,j) = \frac{e^{\phi(h(x_i)) \bigotimes  \psi(h(x_j))}}{\sum_{j=1}^{n}e^{\phi(h(x_i)) \bigotimes  \psi(h(x_j))}}.
\end{equation}
This module is named as `Spatial $m$' in Figure~\ref{Fig:framework}. Here, we compute the correlation score $A_{D}(i,j)$ between node $i$ and node $j$ based on their corresponding representations $h(x_i)$ and $h(x_j)$. The $\bigotimes$ represents matrix multiplication. The $\phi(\cdot)$ and $\psi(\cdot)$ are two projection functions, referred as $conv\_s$ in Figure~\ref{Fig:framework} and can be implemented by channel-wise convolution filters. In this way, the similarity between nodes are captured to build the dynamic graph.

\begin{algorithm}[t]
\caption{CEIM algorithm}\label{algo:CEIM}
\begin{algorithmic}[1]
\Procedure{}{}
\State $\textit{epochs} \gets \text{Max steps of iterations}$
\State $\textit{N} \gets \text{Size of populations in CEIM}$
\State $j \gets \text{Initial iteration step with } \textit{~0}$
\State $i \gets \text{Initial sample index with } \textit{~1}$
\State $\Sigma \gets \text{Initial covariance matrix of CEIM}$
\State $\mu \gets \text{Initial mean of CEIM} $
\State $ \alpha \gets  \text{Initial architecture with}~\mu $
\\

\While {$j < epochs $}
\State {Update network weights $\Theta$ with $\alpha$ fixed.}
\If {$S_{old} ~\text{is not} ~\emptyset $}
\State {Uniformly get ~$r1$ and ~$r2$ in the range [0,1]}.
\While {$i <= N $}
\State {Take a sample $\alpha_{o}^{i}$ from $S_{old}$}.
\If  {~\text{Ineq}~(\ref{eq:old sample})  ~\text{satisfied}}
\State $S_{new} \gets \alpha_{o}^{i}$.
\EndIf
\State {Draw a sample $\alpha_{n}^{i}$ with $\pi \sim \mathcal{N}(\mu, \Sigma)$ }.
\If {~\text{Ineq}~(\ref{eq:new sample})  ~\text{satisfied}} 
\State $S_{new} \gets \alpha_{n}^{i}$.
\EndIf
\State $i = i+1$.
\EndWhile
\EndIf 
\While {$|S_{new}|> N $}
\State {Randomly remove a sample from $S_{new}$}.
\EndWhile
\While {$|S_{new}|< N $}
\State {$S_{new} \gets $ Draw a sample with $\pi$}.
\EndWhile
\State {Evaluate every $\alpha \in S_{new}$ on current network $\Theta$.}
\State {Sort samples by their performances}.
\State {Compute importance weight $\lambda$ by Eq.~(\ref{eq:weight_smaple})}.
\State $ \pi_{old} \gets  \pi $
\State $ S_{old} \gets  S_{new} $
\State {Update $\mu$ and $\Sigma$ by Eq.~(\ref{eq:update_mu}) and Eq.~(\ref{eq:update_sigma})}.
\State $ S_{new} \gets  \emptyset $
\State $ \pi_{new} \gets  \pi $
\State $ \alpha \gets  \mu $
\State $ j = j +1 $
 
\EndWhile
\If {$j == \textit{epochs}$} \Return best $\alpha \in S_{new}.$
\EndIf
\EndProcedure
\end{algorithmic}
\end{algorithm}

\noindent\textbf{Temporal representation Correlation.} Structure correlation definitely contains most intuitive cues for the topology of the graph. However, ignoring the temporal correlation can loss implicit joint correlations. We take an example from NTU RGB+D dataset. Without the temporal information, it is hard to tell a person is to touch his head or just to wave his hand. As from the physical structure perspective, little connection could be captured between head node and hand node during the action of touch head. But there should be a correlation between them in this action. Including the temporal information will make it much easy. Therefore, we introduce two temporal convolutions to extract the temporal information of each node before computing node correlations with Eq~(\ref{eq:gaussain}). In this way, the node interactions between neighbor frames are involved when we calculate the node connections. Note that, temporal representation correlation here is different from temporal attention mechanism, which is to give higher weights to the relative important frames. In contrast, we capture temporal information for a better generation of the spatial graph. To this regard, we also introduce a Gaussian function, as in Eq~(\ref{eq:gaussain}), to compute the node correlation. The functions $\phi(\cdot)$ and $\psi(\cdot)$ are implemented by temporal convolutions, referred as $conv\_t$ in Figure~\ref{Fig:framework} and this module is referred as `Temporal $m$'. It is also worth mentioning that, for the structure correlation, even the T frames of representation are all involved when compute the graph, the interaction is limited to the features from same dimension at the same time step. While our temporal module could involve interactions beyond the same frames.

With these two modules: `Spatial $m$' and `Temporal $m$', it is straightforward to build a spatial-temporal function module for dynamic graph. Thus we build the `Spatio-Temporal $m$' in Figure~\ref{Fig:framework}. Therefore, as in Fig~\ref{Fig:framework}, there are three kinds of modules for dynamic graphs. 

Furthermore, we also want to explore the contribution of each component in the Chebyshev polynomials, and thus benefit from high-order hop connections. As we know, the work in~\cite{kipf2016semi} gives a well-approximation of the spectral filter with the order-one Chebyshev polynomials. Instead, as illustrated in Fig~\ref{Fig:framework}, in our search space, we build Chebyshev polynomials functions with different orders at different layers and let the network determine which order and polynomial components each layer prefers. The function module can be constructed by Eq~(\ref{eq:cheb}), and here the biggest order is $R=4$. Since all the dynamic graphs are normalized, here we also add a normalized one for the order-4 approximation. Therefore, there are totally eight function modules, as illustrated in the Fig~\ref{Fig:framework}, in this search space. 

With these eight modules, we could search for the best architecture. Previous NAS methods would search a single block to reduce the computational burden. However, we argue that different feature layers contain different level of semantic content and thus a layer-specific mechanism is preferred to build a graph. So we search for an entire GCN network instead of a single block. To improve the efficiency, a high computation- and memory-efficient search strategy will be provided. 

Let us formally define the search space first. Here we redefine $X$ as a sequence of graphs. Given a fixed graph $L$ and the feature $h^{k}(X)$ from the $k$-th layer, we extract the output representation $h^{k+1}(X)$ at $k+1$ layer, with the function modules we choose. Inspired by one-shot NAS and DARTS~\cite{liu2018darts}, all the function modules are paralleled and the weighted sum of their outputs are the output $h^{k+1}(X)$, that is
\begin{equation}\label{eq:weighted}
    h^{k+1}(X) = \sum_{i=1}^{M}\frac{\alpha_{k+1,i}}{\sum_{j}^{M} \alpha_{k+1,j}} \mathcal{M}_i(h^{k}(X),L)h^{k}(X)\Theta_k.
\end{equation}
Here, $\Theta_k$ is the network weights for the $k$-th layer. $\mathcal{M}_i$ denotes the $i$-th function module, and $\alpha_{k+1,i}$, which works as the architecture parameter, is its corresponding parameter at the $k+1$ layer. Then the problem here is to search a set of parameters $\alpha\in \mathrm{R}^{K \times M}$ for a network with $K$ layers so that $\alpha$ minimizes the loss $L_{valid}$ on the validation data. That is
\begin{equation}\label{eq:object}
    \alpha^* = \operatorname*{argmin}_{\alpha} L_{valid}({\Theta}(\alpha), \alpha)
\end{equation}
Here, $\Theta$ is the network parameters shared by all sub-networks and it will be learned on the training dataset. Previous works search on a small proxy dataset to evade the expensive computational burden. Instead, here we search directly on the target dataset to avoid introducing extra domain adaption problem.

\subsubsection{GCN search strategy}
Inspired by~\cite{pourchot2018cem}, we propose to search with a high sampling-efficient ES-based method, denoted CEIM. This method explore an optimal architecture by estimating the architecture distribution. Thus it is not limited in a differentiable search space. One could improve the memory-efficiency by only activating one function module at each searching step.

Specifically, this search strategy combines \textbf{C}ross-\textbf{E}ntropy method~\cite{larranaga2001estimation} with \textbf{I}mportance-\textbf{M}ixing (CEIM) to improve the sampling efficiency. In CEIM, architecture parameters $\alpha$ is treated as a population and the distribution of architecture is modeled by a Gaussian distribution. Then CEIM samples a group of architectures and with their performances, important samples are selected to update the architecture distribution. Thus an optimal architecture could be finally sampled from the architecture distribution. 
In total, there are three steps in our CEIM algorithm, sampling populations, selecting populations, and updating architecture distribution. Firstly, we model the architecture distribution with a Gaussian distribution~$\pi \sim \mathcal{N}(\mu, \Sigma)$ and sample $N$ architecture samples $\mathcal{S}_{new}$ as the populations for CEIM, that is $\mathcal{S}_{new}=\{\alpha_n^i\}_{i=1}^{N}$. Secondly, combining $\mathcal{S}_{new}$ with historical selected populations $\mathcal{S}_{old}$, we employ an importance mixing method on all these populations to choose architecture samples. Finally, the newly selected samples are used to update the architecture distribution~$\pi$. 

Here, we detailed the last two steps. Assume samples from previous iteration is $\mathcal{S}_{
old}=\{\alpha_o^i\}_{i=1}^{N}$. In the selecting step, for each population in $\mathcal{S}_{old}$ and $\mathcal{S}_{new}$, we compare its probability density (pd) in both current ($\pi_{new}$) and old ($\pi_{old}$) probability density functions (pdf). Generally, for the old population $\alpha_{o}^i$, we keep it once it is with a bigger pd in the new distribution than that in the old one. That is 
\begin{equation}\label{eq:old sample}
     min(1, \frac{p(\alpha_{o}^{i};\pi_{new})}{p(\alpha_{o}^{i};\pi_{old})})>r1
\end{equation}
Here $r1$ is a threshold randomly got from rang $[0,1]$ and $p(\cdot;\pi)$ is a pdf with specific distribution $\pi$. Likewise, for new sample $\alpha_{n}^i$ drawn from the current distribution, if its pd in the new pdf is bigger than that in the old one, we will also keep it. Therefore, when 
\begin{equation}\label{eq:new sample}
     max(0, 1-\frac{p(\alpha_{n}^{i};\pi_{old})}{p(\alpha_{n}^{i};\pi_{new})})> r2
\end{equation}
we save it. Here $r2$ is another threshold in $[0,1]$.

For the updating step, the samples selected in previous step are used to update mean $\mu$ and convariance $\Sigma$. Before that, the $\Theta$ of the network is updated on the training data with current architecture $\alpha = \mu$. Then, the $\Theta$ is fixed and every selected sample is set as the current architecture. Its corresponding fitness is evaluated on the validation data. With their performances, all the selected samples are sorted. 
Based on the performance order, an importance weight $\lambda_{i}$ is assigned to the $i$-th sample. That is 
\begin{equation}\label{eq:weight_smaple}
\lambda_{i}= \frac{log(1+N)/i}{\sum_{i=1}^{N}log(1+N)/i}. 
\end{equation}

\noindent In this way, the sample with better performance will be given a bigger weight, thus it contributes more to the updating of the distribution. Finally, the weighted samples is applied to update the architecture distribution. That is

\begin{equation}\label{eq:update_mu}
\mu_{new} = \sum_{i=1}^{N}\lambda_{i}\alpha^i,
\end{equation}
\begin{equation}\label{eq:update_sigma}
\Sigma_{new} = \sum_{i=1}^{N}\lambda_{i}(\alpha^i- \mu)^2 + \epsilon\mathcal{I}.
\end{equation}
Here, $\epsilon\mathcal{I}$ is a noise term for better exploring of the neural architecture. Since, in practice, $\Sigma$ is too large to compute and update, here we constrain it to be a diagonal one. Note that in Eq.~(\ref{eq:update_sigma}), different with the original cross-entropy method, which updates $\Sigma$ with the new mean $\mu_{new}$, we use the mean of last iteration to update $\Sigma$ since convariance matrix adaption evolution strategy (CMA-ES) shows it is more efficient~\cite{hansen2016cma}. More details about CEIM please refer to Algorithm.~\ref{algo:CEIM}.

One could improve the memory-efficiency by only activating one function module at each searching step. That means for the output $h^{k+1}(X)$, it can be a single output from the activated module.   
\begin{equation}
    h^{k+1}(X) = \left\{\begin{matrix}
 \mathcal{M}_1(h^{k}(X),L)h^{k}(X)\Theta_{k+1},~p= \frac{\alpha_{k+1,1}}{\sum_{j}^{M} \alpha_{k+1,j}}\\ 
 \cdots \\ 
 \mathcal{M}_M(h^{k}(X),L)h^{k}(X)\Theta_{k+1},~p= \frac{\alpha_{k+1,M}}{\sum_{j}^{M} \alpha_{k+1,j}}\\
\end{matrix}\right.
\end{equation}

\noindent Here, each module is activated by a multinomial distribution with the probability $p\sim \alpha_{k+1,i}$ and $\Theta_{k+1}$ is the activated weight of the ($k+1$)-th layer. In the following section, we will evaluate the proposed method.


\section{Experiments}
To evaluate the performance of our model, we carry out comparative experiments on two large-scale skeleton datasets, NTU RGB+D~\cite{shahroudy2016ntu} and Kenitics-Skeleton~\cite{kay2017kinetics,yan2018stgan}, for action recognition task.

\subsection{Dataset \& Evaluation Metrics} 

\textbf{NTU RGB+D}~~NTU RGB+D is currently the most widely used and the largest multi-modality indoor-captured action recognition dataset. There are RGB videos, depth sequences, infrared videos and 3D skeleton data in it. The skeleton data, which is captured by the Microsoft Kinect v2 camera, is the one we use here. There are totally 56,880 video clips captured from three cameras at different heights with different horizontal angles. These actions cover 60 human action classes including single-actor action, which is from class 1 to 49, and two-actor action, which is from class 50 to 60. There are 25 3D joints coordinates for each actor. We follow the benchmark evaluations in the original work~\cite{shahroudy2016ntu}, which are Cross-subject (CS) and the Cross-view (CV) evaluations.  In the CS evaluation, the training set contains 40,320 videos from 20 subjects, and the rest 16,560 video clips are used for testing. In the CV evaluation, videos captured from camera two and three, contains 37,920 videos, are used in the training and the videos from camera one, contains 18,960 videos, are used for testing. In the comparison, Top-1 accuracy is reported on both of the two benchmarks. 

\noindent \textbf{Kinetics-Skeleton}~~Kinetics-Skeleton is based on the very large scale action dataset Kinetics~\cite{kay2017kinetics}, in which there are approximately 300 000 video clips collected from YouTube. This dataset covers 400 kinds of human actions. However, the original Kinectics dataset has no skeleton data. Yan \etal~employed the open source toolbox OpenPose~\cite{Cao_2017} to estimate the 2D joints location of each frame and then built this huge dataset Kinetics-Skeleton~\cite{yan2018stgan}. For each person, coordinates (X, Y) form 18 joints are estimated. For the frames which contain more than two persons, only the top-2 persons are selected based on the average joint confidence. The released data pads every clips to 300 frames. During comparison, both the Top-1 and Top-5 recognition accuracy are reported since this task is much harder due to its great variety.
\subsection{Implementation details} 
Our framework is implemented on the PyTorch~\cite{paszke2017automatic} and the code is released at here\footnote{The code will be released after the paper published}. To keep consistent with the current state-of-the-art GCN methods~\cite{yan2018stgan,shi2019two}\footnote{Note, these two SOTA methods claims nine blocks in the paper while their released codes shows they implemented the network with ten blocks.}, we introduce ten GCN blocks into our network for both searching and training steps. Each of them is based on the block in Figure~\ref{Fig:framework}. Like the previous works, each block is followed with a temporal convolution with the kernel size $9\times 1$ to capture the temporal information. The first GCN block projects the graph into a feature space with the channel number of 64. Then there will be three layers outputting 64 channels for the outputs. After that, the following three layers double the output channels. And the last three layers have 256 channels for outputs. Just like~\cite{yan2018stgan}, the resnet mechanism is applied on each GCN block. Finally, the extracted features are fed into a fully-connected layer for the final prediction.  

For each GCN block, the spatial modules $conv\_s$ are channel-wise convolution filters and the temporal filters $conv\_t$ are convolution filters with kernel size $9\times 1$ performing along the temporal dimension. During searching, we conduct the experiments on the NTU RGB+D Joint data to find the optimal architecture. We share the same architecture for all the aforementioned datasets to keep consistent with the current state-of-the-art methods. 

For the training process, a stochastic gradient descent (SGD) with Nesterov momentum (0.9) is applied as the optimization algorithm for the network. The cross-entropy loss is selected as the loss function for the recognition task. The weight decay is set to 0.0001 and 0.0006 for searching and training, respectively. For the NTU RGB+D dataset, there are at most two people in each sample of the dataset. If the number of bodies in the sample is less than 2, we pad the second body with 0. The max number of frames in each sample is 300. For samples with less than 300 frames, we repeat the samples until it reaches 300 frames. The learning rate is set as 0.1 and is divided by 10 at the 30th, 45th, and 60th epoch. The training process is ended at the 70th epoch.

\subsection{Architecture search analysis}

\begin{table}[t]\footnotesize
\begin{center}
\vspace{-1em}
\caption{\small{Searched modules at each layer. Here, each row refers to a block layer. There are eight module options, including dynamic graph modules ($\mathcal{M}(S)$, $\mathcal{M}(T)$, $\mathcal{M}(ST)$) with various spatial-temporal cues and Chebshev approximation with different orders ( L, $L^4$n,$L^4$,$L^3$, $L^2$). The modules marked with $\checkmark$ represent modules selected by CEIM.}}
\vspace{2mm}
\label{tab:architect}
\scalebox{0.8}{\begin{tabular}{l| c c c c c c c c }
\toprule
$\mathcal{M}$ & L & $L^4$n &$L^4$ & $L^3$ & $L^2$ & $\mathcal{M}(S)$ & $\mathcal{M}(T)$ & $\mathcal{M}(ST)$\\
\hline 
$K_1$ &  &  &  &  & \checkmark& \checkmark& \checkmark& \checkmark\\
$K_2$ &  &  &  &  & & \checkmark& \checkmark& \checkmark\\
$K_3$ &  &  &  &  & & \checkmark& \checkmark& \checkmark\\
$K_4$ &  &  &  &  & & \checkmark& \checkmark& \checkmark\\
$k_5$ &  &  &  &  & \checkmark& \checkmark& \checkmark& \\
$k_6$ &  &  &  &  & \checkmark& & \checkmark& \\
$k_7$ &  & \checkmark &  &  & \checkmark& \checkmark& \checkmark& \checkmark\\
$k_8$ &  &  &  &  & \checkmark& & \checkmark& \\
$k_9$ &  &  &  &  & \checkmark& & \checkmark& \\
$k_{10}$ &  &  &  &  & & & \checkmark& \\
\bottomrule
\end{tabular}}
\end{center}
\vspace{-2em}

\end{table}

We conduct 70 epochs for searching. For the first 20 epochs, we randomly update each module in the network without evaluating any architectures. After that,  we sample $N=50$ architectures and update the architecture distribution with our CEIM algorithm. When the searching is done, we choose the modules of which the architecture parameters $\alpha > 0.1$. Like ~\cite{shi2019two}, a complementary dynamic graph is added, as we do not involve the weight $\alpha$ for each module in the final implementation. The searched architecture is listed in Table~\ref{tab:architect}. The result shows that different layers prefer different mechanisms to generate graphs, which is consistent with our expectation since high level representation contains more semantic information. Concretely, as in Table~\ref{tab:architect}, the lower layers, like layer $K_1$ to $K_4$, count in all dynamic function modules to capture richer information. For higher layers, the temporal representation correlations are more preferred. It is interesting that temporal graph module~$\mathcal{M}(T)$ is selected through the entire network while the spatial function module~$\mathcal{M}(S)$ is limited to the lower layers, which proves the effectiveness of the proposed temporal function module. For the higher-order connections, we found that 2-order hop connection is much welcomed than any other one. And surprisingly we found the $L$, which encodes the physical structure of the skeleton data, is not involved at any layer. This founding gives us a new inspiration about how to build a GCN.

\subsection{Ablation Study}

Here, we explore the effectiveness of the graph modules and also our searched GCN. Therefore, we perform the following experiments on NTU RGB+D with the benchmark of cross-view. Here we compare with six baselines, which are with different mechanism to build the dynamic graphs. Specifically, the modules used to generate graph are based on: 1) Structure representation correlations (S, here it is 2S-ACGN~\cite{shi2019two}); 2) Temporal representation correlations (Ours(T)); 3) Spatial-Temporal representation correlations(Ours(ST)); 4) Temporal correlations with 4-order Chebyshev approximation (Ours(T+Cheb)); 5) Spatial-Temporal representation correlations with 4-order Chebyshev approximation (Ours(ST+Cheb)); and 6) Combine all aforementioned modules(Ours(S+T+ST+Cheb)).  Inspired by ~\cite{shi2019two}, we evaluate these models on both joints and the bones (the second order information of skeleton joints) data, thus a fusion result from score-level is also reported. For these six methods, the same block is shared through the whole network. Instead, our searched method explores the best modules for different layers. The comparison results are listed in the Table~\ref{tab:ablation study}. It shows that temporal information do helps for GCN (Ours(T) and Ours(ST)) and involving all modules can not make sure a better performance (Ours(S+T+ST+Cheb)). Besides, higher-order also helps for GCN (Ours( +Cheb)).The superior performance of the NAS-based GCN (Ours(NAS)) proves the effectiveness of our method. When compared to the current best result~\cite{shi2019two}, shows in the first row, we improve the accuracy 0.9\%, 1.5\%, and 0.6\% on the Joint, Bone and Combine, respectively. This verifies the effectiveness of our SGCN method.

\begin{table}[t]\footnotesize
\begin{center}
\caption{\small{Ablation Study. Performance comparison on NTU RGB+D with CV evaluation.}}
\vspace{3mm}
\label{tab:ablation study}
\scalebox{0.9}{
\begin{tabular}{l c c c }

\toprule
Methods &Joint(\%) &Bone(\%) &Combine(\%)\\
\hline 

2S-AGCN~\cite{shi2019two}  &93.7 & 93.2 &95.1\\
\midrule
\textbf{Ours(T)}  & 93.8& 93.7&95.1\\
\textbf{Ours(ST)}  & 94.0& 93.8&95.2\\
\textbf{Ours(T+Cheb)}  & 94.0& 93.9&95.2\\
\textbf{Ours(ST+Cheb)}  & 94.2& 93.9&95.3\\
\textbf{Ours(S+T+ST+Cheb)}  & 93.9& 93.6&95.1\\
\midrule
\textbf{Ours(NAS)}  & \textbf{94.6} & \textbf{94.7}& \textbf{95.7}\\

\bottomrule
\end{tabular}}
\end{center}
\vspace{-2em}

\end{table}

\subsection{Comparison with State-of-the-arts (SOTA)} 

To evaluate the performance of our searched model, we compared it with 14 SOTA skeleton-based action recognition approaches, including hand-crafted methods~\cite{hu2015jointly}, CNN-based methods~\cite{kim2017interpretable,liu2017enhanced}, LSTM-based methods~\cite{shahroudy2016ntu,song2017end,zhang2017view,si2018skeleton}, reinforcement learning based method~\cite{tang2018deep}, and current promising GCN-based methods~\cite{li2018spatio,yan2018stgan,li2019spatio,gao2019optimized,li2019actional,shi2019two}, on NTU RGB+D and Kinetics-Skeleton datasets. Table~\ref{tab:NTU} and Table~\ref{tab:Kinetics} show the results on these two datasets, respectively. Here we report the best result after performing the score-level fusion on joints and bones. It can be seen from Table~\ref{tab:NTU} and~\ref{tab:Kinetics} that the searched model achieves the best performance on both of the two datasets in terms of all evaluation metrics. This proves the effectiveness of our method.

\begin{table}[t]\footnotesize
\begin{center}
\caption{ \small{Performance comparison on NTU RGB+D with 14 current state-of-the-art methods. }}

\label{tab:NTU}
\scalebox{0.86}{
\begin{tabular}{lc c l }

\toprule
Methods &CS(\%) &CV(\%) &Conference\\
\hline 
Joint~\cite{hu2015jointly} &60.2 &65.2 &CVPR2015\\
P-LSTM~\cite{shahroudy2016ntu} &62.9 &70.3 &CVPR2016\\
STA-LSTM~\cite{song2017end} &73.4 &81.2 & AAAI2017\\
TCN~\cite{kim2017interpretable} &74.3 &83.1 &CVPRW2017\\
VA-LSTM~\cite{zhang2017view} &79.2 &87.7 &CVPR2017\\
SynCNN~\cite{liu2017enhanced} &80.0 &87.2 & PR2017(Jou.)\\
Deep STGCK~\cite{li2018spatio}& 74.9& 86.3 &AAAI2018\\
ST-GCN~(Yan et al. 2018) &81.5 &88.3 &AAAI2018\\
DPRL~\cite{tang2018deep} &83.5 &89.8 & CVPR2018\\
SR-TSL~\cite{si2018skeleton} & 84.8 &92.4 &ECCV2018\\
STGR-GCN~\cite{li2019spatio}& 86.9 &92.3 &AAAI2019\\
GR-GCN~\cite{gao2019optimized}  &87.5 &94.3& ACMM2019\\
AS-GCN~\cite{li2019actional}  &86.8 &94.2& CVPR2019\\
2S-AGCN~\cite{shi2019two}  &88.5 &95.1& CVPR2019\\
\midrule

\textbf{Ours(Joint+Bone)}  &\textbf{89.4}& \textbf{95.7}&-\\
\bottomrule
\end{tabular}}
\end{center}
\vspace{-2em}

\end{table}

\begin{table}[h]\footnotesize
\begin{center}
\vspace{-1em}
\caption{\small{Performance comparison on Kinetics with eight current state-of-the-art methods.}} 

\label{tab:Kinetics}
\scalebox{0.85}{\begin{tabular}{l c c l }

\toprule
Methods &Top-1(\%) &Top-5(\%) &Conference\\
\hline
Feature~\cite{fernando2015modeling} &14.9 &25.8 & CVPR2015\\
P-LSTM~\cite{shahroudy2016ntu} &16.4 &35.3 &CVPR2016\\
TCN~\cite{kim2017interpretable} &20.3 &40.0 & CVPRW2017\\
ST-GCN~(Yan et al. 2018) &30.7 &52.8 & AAAI2018\\
AS-GCN~\cite{li2019actional}  &34.8 &56.5& CVPR2019\\
2S-AGCN(\textbf{Joint})~\cite{shi2019two}  &35.1 & 57.1 & CVPR2019\\
2S-AGCN(\textbf{Bone})~\cite{shi2019two}  & 33.3 & 55.7& CVPR2019\\
2S-AGCN~\cite{shi2019two}  &36.1 & 58.7& CVPR2019\\
\midrule
\textbf{Ours(Joint)}  & \textbf{35.5}& \textbf{57.9}&-\\
\textbf{Ours(Bone)}  & \textbf{34.9}& \textbf{57.1}&-\\
\textbf{Ours(Joint+Bone)}  & \textbf{37.1}& \textbf{60.1}&-\\
\bottomrule
\end{tabular}}
\end{center}
\vspace{-2em}

\end{table}

\section{Conclusion}
In this paper, we propose to build graph convolutional network for skeleton-based action recognition with neural architecture search (NAS). To enrich the NAS search space, firstly, three dynamic graph generating modules are constructed on the basis of various spatial-temporal correlations of nodes. Secondly, modules with higher-order connections are introduced to enlarge the receptive field of GCN convolution. Besides, we devise a novel search strategy by combining cross-entropy evolution strategy with importance-mixing (CEIM), which is both sampling- and memory-efficient. Based on the proposed NAS method, we explore the optimal GCN architecture in this space for skeleton action recognition. The searched model proves the effectiveness of our temporal-based dynamic graph module. Comprehensive experiments on two very large-scale datasets show its overwhelming performance when compared to the state-of-the-art approaches.

\bibliographystyle{aaai}
\bibliography{aaai19-main}

\end{document}